\documentclass[twoside]{llncs}

\usepackage{graphicx}
\usepackage{amsmath}
\usepackage{array}
\usepackage[colorlinks=true,%
            linkcolor=blue,%
            citecolor=blue,%
            urlcolor=blue]{hyperref}           % for hyperlinks

\usepackage{fancyhdr}           % For LaTeX2.09 you should specify {fancyhdr}
\usepackage{ifthen}

\newcommand\settitle[2][]{%
 \title{#2}
 \ifthenelse{\equal{#1}{}}%
  {\fancyhead[RO]{\nouppercase #2 \qquad \thepage}}%
  {\fancyhead[RO]{\nouppercase #1 \qquad \thepage}}%
}

\newcommand\setauthors[2]{%
 \author{#2}
  {\fancyhead[LE]{\thepage \qquad \nouppercase #1}}%
}

\fancypagestyle{plain}{%
\fancyhf{}
}

\def\keywordsname{Keywords.}
\newenvironment{keywords}{%
      \list{}{\advance\topsep by-0.50cm\relax\small
      \leftmargin=1cm
      \labelwidth=1cm%\z@
      \listparindent=1cm%\z@
      \itemindent\listparindent
      \rightmargin\leftmargin}\item[\hskip\labelsep
                                    \bfseries\keywordsname]}
    {\endlist}
\newcommand{\R}{I\!R}             % real number space
\begin{document}

%------------------------------------------------------------------------
% \settitle :
% the optional argument [...] should be only one line
% if the optional argument [...] exists, it will be used for the header
% otherwise the obligatory argument {...} will be used for header
% the obligatory argument is used as FULL title

\settitle[Complexity of Optimized Crossover]
         {On Complexity of Optimized Crossover for Binary Representations}
% title for special papers: Category e.g. Summary, Discussion, Open Problems, ...
% choose ''Dagstuhl Seminar''or ''Perspectives Workshop''
%\settitle[seminar-number -- Paper Category]
%         {{\large seminar-number -- Paper Category} \\
%          Please Fill in Your own FULL Title \\
%         %\large{--- Perspectives Workshop ---}}
%          {\large --- Dagstuhl Seminar ---}}

%------------------------------------------------------------------------
% \setauthors :
% the first argument will be used for the header; it should be only one line
% the second argument is the FULL list of authors

% Warning: Do not use \and together with \usepackage{hyperref}
\setauthors{Anton Eremeev}
           {Anton Eremeev\inst{1}}

\institute{$^1$ Omsk Branch of Sobolev Institute of Mathematics, \\
Laboratory of Discrete Optimization\\
 644099, Omsk, 13, Pevtsov str., Russia\\
\email{eremeev@ofim.oscsbras.ru} }

%------------------------------------------------------------------------
\date{}
\maketitle

\thispagestyle{plain}

%------------------------------------------------------------------------
%------------------------------------------------------------------------
\begin{abstract}
We consider the computational complexity of producing the best
possible offspring in a crossover, given two solutions of the
parents. The crossover operators are studied on the class of
Boolean linear programming problems, where the Boolean vector of
variables is used as the solution representation. By means of
efficient reductions of the optimized gene transmitting crossover
problems (OGTC) we show the polynomial solvability of the OGTC for
the maximum weight set packing problem, the minimum weight set
partition problem and for one of the versions of the simple plant
location problem. We study a connection between the OGTC for
linear Boolean programming problem and the maximum weight
independent set problem on 2-colorable hypergraph and prove the
NP-hardness of several special cases of the OGTC problem in
Boolean linear programming.
\end{abstract}

%------------------------------------------------------------------------
%------------------------------------------------------------------------
% old keywords command
% \keywords{Dagstuhl Seminar Proceedings, style, template}

%------------------------------------------------------------------------
%------------------------------------------------------------------------
\section{Introduction \label{intro}}
In this paper, the computational complexity of producing the best
possible offspring in a crossover, complying with the principle of
respect (see e.g.~\cite{R91}) is considered. The focus is on the
gene transmitting crossover operators, where all alleles present
in a child are transmitted from its parents. These operators are
studied on the Boolean linear programming problems, and in most of
the cases the Boolean vector of variables is used as the solution
representation.

One of the well-known approaches to analysis of the genetic
algorithms (GA) is based on the schemata, i.e. the sets of
solutions in binary search space, where certain coordinates are
fixed to zero or one. Each evaluation of a genotype in a GA can be
regarded as a statistical sampling event for each of~$2^n$
schemata, containing this genotype~\cite{Holl}. This parallelism
can be used to explain why the schemata that are fitter than
average of the current population are likely to increase their
presence (e.g. in Schema Theorem in the case of Simple Genetic
Algorithm).

An important task is to develop the recombination operators that
efficiently manipulate the genotypes (and schemata) producing
"good"\ offspring chromosomes for the new sampling points. An
alternative to random sampling is to produce the {\em best
possible} offspring, respecting the main principles of schemata
recombination. One may expect that such a synergy of the
randomized evolutionary search with the optimal offspring
construction may lead to more reliable information on "potential"\
of the schemata represented by both of the parent genotypes and
faster improvement of solutions quality as a function of the
iterations number. The results in~\cite{AOT97,BN98,BDE,Reeves94}
and other works provide an experimental support to this reasoning.

The first examples of polynomially solvable optimized crossover
problems for NP-hard optimization problems may be found in the
works of C.C.~Aggarwal, J.B.~Orlin and R.P.~Tai~\cite{AOT97} and
E.~Balas and W.~Niehaus~\cite{BN98}, where the optimized crossover
operators were developed and implemented in GAs for the maximum
independent set and the maximum clique problems. We take these
operators as a starting point in Section~\ref{PolySolve}.

By the means of efficient reductions between the optimized gene
transmitting crossover problems~(OGTC) we show the polynomial
solvability of the OGTC for the maximum weight set packing
problem, the minimum weight set partition problem and for one of
the versions of the simple plant location problem. In the present
paper, all of these problems are considered as special cases of
the Boolean linear programming problem: maximize
\begin{equation}\label{goal}
   f(x) = \sum_{j=1}^{n} c_{j}x_{j},
\end{equation}
subject to
\begin{equation} \label{ineq}
\sum_{j=1}^{n} a_{ij} x_{j} \leq b_i, \quad i=1,\dots,m,
\end{equation}
\begin{equation} \label{bools}
x_j \in \{0,1\}, \quad j=1,\dots,n.
\end{equation}
Here $x \in \{0,1\}^n$ is the vector of Boolean variables, and the
input data $c_{j}$, $a_{ij}$, $b_i$ are all integer (arbitrary in
sign). Obviously, this formulation also covers the problems where
the inequality sign "$\le$"\ in~(\ref{ineq}) is replaced by
"$\ge$"\ or "$=$"\ for some or all of indices $i$. The
minimization problems are covered by negation of the goal
function. In what follows, we will use a more compact notation for
problem (\ref{goal})--(\ref{bools}):
$$
\max\left\{cx : Ax \le b, x \in \{0,1\}^n\right\}.
$$

In Section~\ref{HyperGraph} we consider several NP-hard cases of
the OGTC problem. The OGTC for linear Boolean programming problem
with logarithmically upper-bounded number of non-zero coefficients
per constraint is shown to be efficiently reducible to the maximum
weight independent set problem on 2-colorable hypergraph with
2-coloring given as an input. Both of these OGTC problems turn out
to be NP-hard, as well as the OGTC for the set covering problem
with binary representation of solutions.

%------------------------------------------------------------------------
%------------------------------------------------------------------------

\section{Optimized Recombination and Principle of Respect \label{PolySolve}}
We will use the standard notation to define schemata. Each schema
is identified by its indicator vector $\xi \in \{0,1,*\}^n,$
implying the set of genotypes
$$
\left\{x\in\{0,1\}^n : x_j=\xi_j \mbox{ for all } j \mbox{ such
that } \xi_j =0  \mbox{ or } \xi_j =1\right\}
$$
attributed to this schema (the elements~$x$ are also called the
instances of the schema).

Suppose, a set of schemata on Boolean genotypes is defined:
$\Xi\subseteq \{0,1,*\}^n.$ Analogously to
N.J.~Radcliffe~\cite{R91}, we can require the optimized crossover
on Boolean strings to obey the principle of {\em respect}:
crossing two instances of any schema from $\Xi$ should produce an
instance of that schema. In the case of Boolean genotypes and
$\Xi=\{0,1,*\}^n$ this automatically implies the {\em gene
transmission property}: all alleles present in the child are to be
transmitted from its parents.

In this paper, we will not consider the principle of {\em
ergodicity} which requires that it should be possible, through a
finite sequence of applications of the genetic operators, to
access any point in the search space given any initial population.
Often this property may be ensured by the means of mutation
operators but they are beyond the scope of the paper. Besides
that, we shall not discuss the principle of {\em proper
assortment}: given instances of two compatible schemata, it should
be possible to cross them to produce a child which is an instance
of both schemata. This principle appears to be irrelevant to the
optimized crossover.

In what follows we shall use the standard definition of
NP~optimization problem (see e.g.~\cite{AP95}). By $\{0,1\}^*$ we
denote the set of all strings with symbols from~$\{0,1\}$ and
arbitrary string length.

\begin{definition}
An $NP$~optimization problem $\Pi$ is a triple
${\Pi=(I,Sol,f_X)}$, where $I \subseteq \{0,1\}^*$
is the set of instances of~$\Pi$ and:\\

1. $I$ is recognizable in polynomial time (through this paper the
term polynomial time implies the running time bounded by
a polynomial on length of input instance encoding $|X|, X \in I$).\\

2. Given an instance $X \in I$, $Sol(X)\subseteq \{0,1\}^{n(X)}$
is the set of feasible solutions of $X$. Given $X$ and $x$, the
decision whether $x\in Sol(X)$ may be done in polynomial time,
and $n(X) \le h(|X|)$ for some polynomial $h$.\\

3. Given an instance $X \in I$ and $x\in Sol(X)$, $f_X: Sol(X) \to
\R$ is the objective function (computable in polynomial time) to
be maximized if $\ \Pi$ is an NP~maximization problem or to be
minimized if $\ \Pi$ is an NP~minimization problem.
\end{definition}

In this definition $n(X)$ stands for the dimension of Boolean
space of solutions of problem instance $X$. In case different
solutions have different length of encoding, $n(X)$ equals the
size of the longest solution. If some solutions are shorter
than~$n(X)$, the remaining positions are assumed to have zero
values. In what follows, we will explicitly indicate the method of
solutions representation for each problem since it is crucial for
the crossover operator.

\begin{definition}
For an NP~maximization problem $\Pi_{\max}$ the optimized gene
transmitting crossover problem (OGTC) is formulated the following
way.

Given an instance $X$ of \ $\Pi_{\max}$ and two {\em parent
solutions} $p^1,p^2 \in Sol(X)$, find an {\em offspring solution}
$x \in Sol(X)$, such that

(a) $x_j =p^1_j$ or $x_j =p^2_j$ for each $j=1,\dots,n(X)$, and

(b) for any $x'\in Sol(X)$ such that $x'_j =p^1_j$ or $x'_j
=p^2_j$ for all $j = 1,\dots,n(X)$, holds $f_X(x) \ge f_X(x').$
\end{definition}

A definition of the OGTC problem in the case of NP~minimization
problem is formulated analogously, with the modification of
condition~(b):

{\em (b') for any $x'\in Sol(X)$, such that $x'_j =p^1_j$ or $x'_j
=p^2_j$ for all $j = 1,\dots,n(X)$, holds $f_X(x) \le f_X(x').$}

In what follows, we denote the set of coordinates, where the
parent solutions have different values, by $D(p^1,p^2)=\{j: p^1_j
\ne p^2_j\}.$

The optimized crossover problem could be formulated with a
requirement to respect some other set of schemata, rather than
$\{0,1,*\}^n$. For example, the set of schemata $\Xi=\{0,*\}^n$
defines the optimized crossover operator used in~\cite{Er99} for
the set covering problem. For such $\Xi$ condition~(a) is
substituted by $x_j \le p^1_j+p^2_j$ for all~$j$. The crossover
subproblems of this type will have a greater dimension than the
OGTC problem and they do not possess the gene transmission
property. In what follows, we will concentrate only on the OGTC
problems.

As the first examples of efficiently solvable OGTC problems we
will consider the following three well-known problems. Given a
graph $G=(V,E)$ with vertex weights $w(v), \ v \in V$,

\begin{itemize}
\item the maximum weight independent set problem asks for a subset
${S\subseteq V}$, such that each ${e \in E}$ has at least one
endpoint outside~$S$ (i.e. $S$ is an independent set) and the
weigth $\sum_{v \in S} w_v$ of $S$ is maximized;

\item the maximum weight clique problem asks for a maximum weight
subset ${Q\subseteq V}$, such that any two vertices~$u,v$ in~$Q$
are adjacent;

\item the minimum weight vertex cover problem asks for a minimum
weight subset ${C\subseteq V}$, such that any edge ${e \in E}$ is
incident at least to one of the vertices in~$C$.
\end{itemize}

Suppose, all vertices of graph $G$ are ordered. We will consider
these three problems using the standard binary representation of
solutions by the indicator vectors, assuming $n=|V|$ and $x_j=1$
iff vertex~$v_j$ belongs to the represented subset.
Proposition~\ref{AOT} below immediately follows from the results
of E.~Balas and ~W.~Niehaus~\cite{BN96} for the unweighted case
and~\cite{BN98} for the weighted case.

\begin{proposition}\label{AOT} The OGTC problems for
the maximum weight independent set problem, the maximum weight
clique problem and the minimum weight vertex cover problem are
solvable in polynomial time in the case of standard binary
representation.
\end{proposition}

The efficient solution method for these problems is based on a
reduction to the maximum flow problem in a bipartite graph induced
by union of the parent solutions or their complements (in the
unweighted case the maximum matching problem is applicable as
well). The algorithm of A.V.~Karzanov allows to solve this problem
in $O(n^3)$ steps, but if all weights are equal, then its time
complexity reduces to $O(n^{2.5})$ -- see e.g.~\cite{PS98} . The
algorithm of A.~Goldberg and R.~Tarjan~\cite{GT88} has a better
performance if the number of edges in the subgraph is considered.

The usual approach to spreading a class of polynomially solvable
(or intractable) problems consists in building the chains of
efficient problem reductions. The next proposition serves this
purpose.

\begin{proposition}\label{reduction0}
Let ${\Pi_1=(I_1,Sol_1,f_{X})}$ and ${\Pi_2=(I_2,Sol_2,g_{Y})}$ be
both NP~maximization problems and $Sol_1(X)\subseteq
\{0,1\}^{n_1(X)}$ and $Sol_2(Y)\subseteq \{0,1\}^{n_2(Y)}$.
Suppose the OGTC is solvable in polynomial time for~$\Pi_2$ and
the following three polynomially computable functions exist:

$\alpha: I_1 \to I_2$,

$\beta\  : \ Sol_1(X) \to Sol_2(\alpha(X))$, bijection with the
inverse mapping

$\beta^{-1} \ :\ Sol_2(\alpha(X)) \to Sol_1(X)$,\\
and

(i) For any $x,x' \in Sol_1(X)$ such that $f_{X}(x) < f_{X}(x')$,
holds $g_{\alpha(X)}(\beta(x)) < g_{\alpha(X)}(\beta(x')).$

(ii) for any $j=1,\dots,n_1(X)$, such that ${x_j}$ is not constant
on~$Sol_1(X)$, there exists such $k(j)$ that either
$\beta(x)_{k(j)}=x_j$ for all $x \in Sol_1(X)$, or
$\beta(x)_{k(j)} = 1-x_j$ for all $x \in Sol_1(X)$.

(iii) for any $k=1,\dots,n_2(X)$ exists such~$j(k)$ that
$\beta(x)_{k}$ is a function of~$x_{j(k)}$ on $Sol_1(X)$.

Then the OGTC problem is polynomially solvable for~$\Pi_1$.
\end{proposition}

{\bf Proof.} Suppose, an instance~$X$ of problem~$\Pi_1$ and two
parent solutions $p^1,p^2 \in Sol_1(X)$ are given. Consider two
feasible solutions $q^1=\beta(p^1)$, $q^2=\beta(p^2)$ in
$Sol_2(\alpha(X))$. Let us apply an efficient algorithm to solve
the OGTC problem for the instance $\alpha(X)\in \Pi_2$ with parent
solutions~$q^1, q^2$ (such an algorithm exists by the assumption).
The obtained solution $y\in Sol_2(\alpha(X))$ can be transformed
in polynomial time into $z=\beta^{-1}(y) \in Sol_1(X)$.

Note that for all $j \not \in D(p^1,p^2)$ holds $z_j=p^1_j=p^2_j$.
Indeed, consider the case where in the condition~(ii) for~$j$ we
have $\beta(x)_{k(j)}=x_{j}, \ {x \in Sol_1(X)}$. Hence,
$z_j=y_{k(j)}$. Now $y_{k(j)} = q^1_{k(j)}$ by definition of the
OGTC problem, since $q^1_{k(j)}=p^1_j=p^2_j=q^2_{k(j)}$, so
$z_j=q^1_{k(j)}=p^1_j=p^2_j.$

The case $\beta(x)_{k(j)}=1-x_{j}, \ {x \in Sol_1(X)}$ is treated
analogously. Finally, the case of constant~$x_j$ over $Sol_1(X)$
is trivial since $z,p^1,p^2 \in Sol_1(X)$.

To prove the optimality of~$z$ in OGTC problem for~$\Pi_1$ we will
assume by contradiction that there exists $\zeta \in Sol_1(X)$
such that $\zeta_j=p^1_j=p^2_j$ for all $j \not\in D(p^1,p^2)$ and
$f_X(\zeta)>f_X(z)$. Then $g_{\alpha(X)}(\beta(\zeta)) >
g_{\alpha(X)}(\beta(z))=g_{\alpha(X)}(y)$. But $\beta(\zeta)$
coincides with~$y$ in all coordinates $k\not\in D(q^1,q^2)$
according to condition~(iii), thus~$y$ is not an optimal solution
to the OGTC problem
for~$\alpha(X)$, which is a contradiction. Q.E.D.\\

Note that if ${\Pi_1}$ or~${\Pi_2}$ or both of them are
NP~minimization problems then the statement of
Proposition~\ref{reduction0} is applicable with a reversed
inequality sign in one or both of the inequalities of
condition~(i).\\

Let us apply Proposition~\ref{reduction0} to obtain an efficient
OGTC algorithm for the {\em set packing problem:}
\begin{equation}\label{SPP}
\max\left\{f_{pack}(x)=cx : Ax \le e, x \in \{0,1\}^n\right\},
\end{equation}
where $A$ is a given $(m\times n)$-matrix of zeros and ones and
$e$ is an $m$-vector of ones. The transformation $\alpha$ to the
maximum weight independent set problem with standard binary
representation consists in building a graph on a set of vertices
$v_1,\dots,v_n$ with weights $c_1,\dots,c_n$. Each pair of
vertices $v_j, v_k$ is connected by an edge iff $j$ and $k$ both
belong at least to one of the subsets $N_i = \{j: a_{ij} =1\}.$ In
this case $\beta$ is an identical mapping. Application of
Proposition~\ref{reduction0} leads to

\begin{corollary}\label{setpack} The OGTC problem is polynomially
solvable for the maximum weight set packing problem
{\em(\ref{SPP})} if the solutions are represented by vectors $x
\in \{0,1\}^n$.
\end{corollary}

In some reductions of NP optimization problems the set of feasible
solutions of the original instance corresponds to a subset of
"high-quality"\ feasible solutions in the transformed formulation.
In order to include the reductions of this type into
consideration, we will define the subset of "high-quality"\
feasible solutions for an NP maximization problem as
$$
Sol^X_2(\alpha(X))=\left\{y\in Sol_2(\alpha(X)) : g(y) \ge
\min_{x\in Sol_1(X)} g(\beta(x))\right\},
$$
and for an NP minimization problem
$$
Sol^X_2(\alpha(X))=\left\{y\in Sol_2(\alpha(X)) : g(y) \le
\max_{x\in Sol_1(X)} g(\beta(x))\right\}.
$$
A slight modification of the proof of Proposition~\ref{reduction0}
yields the following

\begin{proposition}\label{reduction1}
The statement of Proposition~\ref{reduction0} also holds if
$Sol_2(\alpha(X))$ is substituted by $Sol^X_2(\alpha(X))$
everywhere in its formulation, implying that $\beta$ is a
bijection from $Sol_1(X)$ to $Sol^X_2(\alpha(X))$.
\end{proposition}

Now we can prove the polynomial solvability of the next two
problems in the Boolean linear programming formulations.

\begin{itemize}
\item The {\em minimum weight set partition problem:}
\begin{equation}\label{SPartP}
\min\left\{f_{part}(x)=cx : Ax = e, x \in \{0,1\}^n\right\},
\end{equation}
where $A$ is a given $(m\times n)$-matrix of zeros and ones.

\item The {\em simple plant location problem:} minimize
\begin{equation}\label{SPPLgoal}
   f_{sppl}(x,y) = \sum_{k=1}^K \sum_{\ell=1}^{L} c_{k\ell}x_{k\ell} +
    \sum_{k=1}^K C_{k}y_{k},
\end{equation}
subject to
\begin{equation}\label{SPPLserve}
\sum_{k=1}^{K} x_{k\ell} = 1, \quad \ell=1,\dots,L,
\end{equation}
\begin{equation}\label{SPPLopen}
y_k \ge x_{k\ell}, \quad k=1,\dots,K, \ \ell=1,\dots,L,
\end{equation}
\begin{equation}\label{SPPLbool}
x_{k\ell} \in \{0,1\}, \ y_{k}\in \{0,1\}, \quad k=1,\dots,K, \
\ell=1,\dots,L.
\end{equation}
Here $x\in \{0,1\}^{KL},y \in \{0,1\}^{K}$ are the vectors of
Boolean variables. The costs~$c_{k\ell}$, $C_{k}$ are nonnegative
and integer.
\end{itemize}

\begin{corollary}\label{setpart_sppl} The OGTC problem is polynomially
solvable for

(i) the minimum weight set partition problem~(\ref{SPartP}) if the
solutions are represented by vectors $x \in \{0,1\}^n$ and

(ii) the simple plant location problem, if the solutions are
represented by couples of vectors $(x,y)$, $x \in \{0,1\}^{KL}$,
$y \in \{0,1\}^{K}$.
\end{corollary}

{\bf Proof.} For both problems we will use the well-known
transformations~\cite{KP83}.

(i) Let us denote the minimum weight set partition problem
by~$\Pi_1$. The input of its OGTC problem consists of an instance
$X\in I_1$ and two parent solutions, thus $Sol_1(X)\ne \emptyset$
and $X$ can be transformed into an instance $\alpha(X)$ of the
following NP~minimization problem~$\Pi_2$ (see the details in
derivation of transformation~T5 in~\cite{KP83}:
$$
\min\left\{g(x)=\sum_{j=1}^n\left(c_j-\lambda\sum_{i=1}^m
a_{ij}\right)x_j : Ax \le e, x \in \{0,1\}^n\right\},
$$
where $\lambda>2\sum_{j=1}^n |c_j|$ is a sufficiently large
constant. We will assume that~$\beta$ is an identical mapping.
Then each feasible solution $x$ of the set partition problem
becomes a "high quality"\ feasible solution to problem~$\Pi_2$
with a goal function value $g(x)=f_{part}(x)-\lambda m < -\lambda
(m-1/2).$ At the same time, if a vector~$x'$ is feasible for
problem~$\Pi_2$ but infeasible in the set partition problem, it
will have a goal function value $g(x')=f_{part}(x')-\lambda
(m-k),$ where $k$ is the number of constraints $\sum_{j=1}^n
a_{ij} x_j=1,$ violated by~$x'$. In other words, $\beta$ is a
bijection from $Sol_1(X)$ to
$$
Sol^X_2(\alpha(X))=\{x\in Sol_2(\alpha(X)) : g(x) < \lambda
(m-1/2)\}.
$$
Note that solving the OGTC for NP~minimization problem~$\Pi_2$ is
equivalent to solving the OGTC for the set packing problem with
the maximization criterion $-g(x)$ and the same set of
constraints. This problem can be solved in polynomial time by
Corollary~\ref{setpack}. Thus, application of
Proposition~\ref{reduction1} completes the proof of part~(i).

(ii) Let~$\Pi'_1$ be the simple plant location problem. We will
use the transformation T2  \hspace{0.1em} from~\cite{KP83} for our
mapping
 \hspace{0.1em} $\alpha(X)$, \hspace{0.2em} which reduces \hspace{0.2em}
(\ref{SPPLgoal})--(\ref{SPPLbool})  \hspace{0.3em} to~the
following~NP~minimization problem~$\Pi'_2$: minimize
\begin{equation}\label{P2goal}
   g'(x,y) = \sum_{k=1}^K \sum_{\ell=1}^{L} (c_{k\ell} - \lambda)x_{k\ell}
   -
    \sum_{k=1}^K C_{k}\overline{y}_{k},
\end{equation}
subject to
\begin{equation}\label{P2serve}
\sum_{k=1}^{K} x_{k\ell} \le 1, \quad \ell=1,\dots,L,
\end{equation}
\begin{equation}\label{P2open}
\overline{y}_k + x_{k\ell} \le 1, \quad k=1,\dots,K, \
\ell=1,\dots,L,
\end{equation}
\begin{equation}\label{P2bool}
x_{k\ell} \in \{0,1\}, \ \overline{y}_{k}\in \{0,1\}, \quad
k=1,\dots,K, \ \ell=1,\dots,L,
\end{equation}

where $x\in \{0,1\}^{KL},\overline{y} \in \{0,1\}^{K}$ are the
vectors of variables and
$$
\lambda>\max_{\ell=1,\dots,L}\left\{\min_{k=1,\dots,K}
\{C_k+c_{k\ell}\}\right\}
$$
is a sufficiently large constant. We will assume that~$\beta$ maps
identically all variables $x_{k\ell}$ and transforms the rest of
the variables as $\overline{y}_k = 1-y_k, \ k=1,\dots,K$. Then
each feasible solution $(x,y)$ of the simple plant location
problem becomes a "high quality"\ feasible solution to
problem~$\Pi'_2$ with a goal function value
$g'(x,\overline{y})=f_{sppl}(x,y)-\lambda L - C_{sum} \le -\lambda
L - C_{sum},$ where $C_{sum} =\sum_{k=1}^K C_k$. At the same time
if a pair of vectors~$(x',\overline{y})$ is feasible for
problem~$\Pi'_2$ but $(x',y)$ is infeasible in the simple plant
location problem, then $g'(x',\overline{y})=f_{sppl}(x',y)-\lambda
(L-k) - C_{sum},$ where $k$ is the number of
constraints~(\ref{SPPLserve}), violated by~$(x',y)$. Solving the
OGTC for NP~minimization problem~$\Pi'_2$ is equivalent to solving
the OGTC for the set packing problem with the maximization
criterion $-g'(x,\overline{y})$ and the same set of constraints.
This can be done in polynomial time by Corollary~\ref{setpack},
thus Proposition~\ref{reduction1} gives an efficient algorithm
solving the OGTC for $\Pi'_1$. Q.E.D.\\

If a vector $y\in \{0,1\}^K$ is fixed, then the best possible
solution to the simple plant location problem with this~$y$ can be
easily constructed: for each~$\ell$ one has to assign one of the
variables $x_{k\ell}=1$, so that $c_{k\ell}\le c_{k'\ell}$ for all
such $k'$ that $y_{k'}=1$. Then it suffices to specify just a
vector~$y$ to represent a tentative solution to this problem. It
is easy to see that it is impossible to construct some non-optimal
feasible solutions to problem~(\ref{SPPLgoal})--(\ref{SPPLbool})
this way. Strictly speaking, the representation given by the
vector~$y$ applies to another NP-minimization problem with a
reduced set of feasible solutions. In the next section it will be
proven that the OGTC for this version of the simple plant location
problem is NP-hard.

%------------------------------------------------------------------------
\section{Some NP-hard Cases of Optimized Crossover Problems \label{HyperGraph}}

The starting point of all reductions in the previous section was
Proposition~\ref{AOT} based on efficient reduction of some OGTC
problems to the maximum weight independent set problem in a
bipartite graph. In order to generalize this approach now we will
move from ordinary graphs to hypergraphs. A hypergraph $H=(V,E)$
is given by a finite nonempty set of vertices~$V$ and a set of
edges~$E$, where each edge~$e\in E$ is a subset of~$V$. A subset
$S\subseteq V$ is called {\em independent} if none of the edges $e
\in E$ is a subset of~$S$. The maximum weight independent set
problem on hypergraph $H=(V,E)$ with integer vertex weights $w_v,
\ v \in V$ asks for an independent set $S$ with maximum weight
$\sum_{v \in S} w_v$. A generalization of the case of bipartite
graph is the case of {\em 2-colorable hypergraph}: there exists a
partition of the vertex set~$V$ into two disjoint independent
subsets $C_1$ and $C_2$ (the partition $V=C_1\cup C_2$, $C_1 \cap
C_2=\emptyset$ is called a 2-coloring of $H$ and $C_1, C_2$ are
the color classes).

Let us denote the set of non-zero elements in constraint~$i$
by~$N_i$:
$$
N_i = \{j: a_{ij} \ne 0\}.
$$

\begin{proposition}\label{hyper2}
Suppose, $|N_i|=O(\ln n)$ for all $i=1,\dots,m$. Then the OGTC for
Boolean linear programming problem is polynomially reducible to
the maximum weight independent set problem on 2-colorable
hypergraph with 2-coloring given in the input.
\end{proposition}

{\bf Proof.} Given an instance of the Boolean programming problem
with parent solutions~$p^1$ and~$p^2$, let us denote
$d=|D(p^1,p^2)|$ and construct a hypergraph~$H$ on~$2d$ vertices,
assigning each variable $x_j, j\in D(p^1,p^2)$ a couple of
vertices $v_j, v_{n+j}$. In order to model each of the linear
constraints for $i=1,\dots,m$ one can enumerate all combinations
$x^{ik} \in \{0,1\}^{|N_i\cap D(p^1,p^2)|}$ of the Boolean
variables from $D(p^1,p^2),$ involved in this constraint. For each
combination~$k$ violating the constraint
$$
\sum_{j\in N_i\cap D(p^1,p^2)} a_{ij} x^{ik}_j
 +
\sum_{j\not\in D(p^1,p^2)} a_{ij} p^{1}_j \le b_i
$$
we add an edge
$$
\{v_j: x^{ik}_j=1, \ j \in N_i\cap D(p^1,p^2)\} \cup
 \{v_{j+n}: x^{ik}_j=0, \ j \in N_i\cap D(p^1,p^2)\}
$$
into the hypergraph. Besides that, we add~$d$ edges
$\{v_j,v_{n+j}\}, j\in D(p_1,p_2)$, to guarantee that both $v_j$
and~$v_{n+j}$ can not enter in any independent set together.

If~$x$ is a feasible solution to the OGTC problem, then
$S(x)=\{v_j: x_j=1\} \cup \{v_{j+n}: x_j=0\}$ is independent
in~$H$. Given a set of vertices~$S$, we can construct the
corresponding vector~$x(S)$ with $x(S)_j=1$ iff $v_j \in S, j\in
D(p^1,p^2)$ or $p^1_j=p^2_j=1$. Then for each independent set~$S$
of~$d$ vertices, $x(S)$ is feasible in the Boolean linear
programming problem.

The hypergraph vertices are given the following weights:
$w_j=c_j+\lambda, \ w_{n+j}=\lambda, j\in D(p^1,p^2),$ where
$\lambda>2\sum_{j\in D(p_1,p_2)} |c_j|$ is a sufficiently large
constant.

Now each maximum weight independent set~$S^*$ contains
either~$v_j$ or~$v_{n+j}$ for any $j\in D(p^1,p^2)$. Indeed, there
must exist a feasible solution to the OGTC problem and it
corresponds to an independent set of weight at least~$\lambda d$.
However, if an independent set does not contain neither~$v_j$
nor~$v_{n+j}$ then its weight is at most $\lambda d - \lambda/2$.

So, optimal~$S^*$ corresponds to a feasible vector~$x(S^*)$ with
the goal function value
$$
cx(S^*)=
 \sum_{j\in S^*, \ j\le n} c_{j}
 +
 \sum_{j\not\in D(p^1,p^2)} c_{j} p^{1}_j
 =
 w(S^*)-\lambda d
 +
 \sum_{j\not\in D(p^1,p^2)} c_{j} p^{1}_j.
$$
Under the inverse mapping~$S(x)$ any feasible vector~$x$ yields an
independent set of weight $cx+\lambda d - \sum_{j\not\in
D(p^1,p^2)} c_{j} p^{1}_j$, so~$x(S^*)$ must be an
optimal solution to the OGTC problem as well. Q.E.D.\\

Note that if the Boolean linear programming problem is a
multidimensional knapsack problem
\begin{equation}\label{knapsack}
\max\left\{cx : Ax \le b, x \in \{0,1\}^n\right\}
\end{equation}
with all $a_{ij} \ge 0$, then the above reduction may be
simplified. One can exclude all vertices $v_{n+j}$ and edges
$\{v_j,v_{n+j}\}$, $j\ge 1$ from~$H$, and repeat the whole proof
of Proposition~\ref{hyper2} with $\lambda=0$. The only difference
is that the feasible solutions of OGTC problem now correspond to
arbitrary independent sets, not only those of size~$d$ and the
maximum weight independent sets do not necessarily contain
either~$v_j$ or~$v_{n+j}$ for any $j\in D(p^1,p^2)$. This
simplified reduction is identical to the one in
Proposition~\ref{AOT} if~$A$ is an incidence matrix of the
ordinary graph~$G$ given for the maximum weight independent set
problem and $b=e$. Polynomial solvability of the maximum weight
independent set problem on bipartite ordinary graphs yields the
polynomial solvability the OGTC for the Boolean multidimensional
knapsack problem where $|N_i|=2, i=1,\dots,m$.

Providing a 2-coloring together with the hypergraph may be
important in the cases, where the 2-coloring is useful for finding
the maximum weight independent set. For example in the special
case where each edge consists of~4 vertices, finding a 2-coloring
for a 2-colorable hypergraph is NP-hard~\cite{GHS02}. However, the
next proposition indicates that in the general case of maximum
independent set problem on 2-colorable hypergraphs, providing a
2-coloring does not help a lot.

\begin{proposition}
Finding maximum size independent set in a hypergraph with all
edges of size~3 is NP-hard even if a 2-coloring is given.
\end{proposition}
{\bf Proof.} Let us construct a reduction from the maximum size
independent set problem on ordinary graph to our problem. Given a
graph~$G=(V,E)$ with the set of vertices $V=\{v_1,\dots,v_{n}\}$,
consider a hypergraph~$H=(V',E')$ on the set of vertices
$V'=\{v_1,\dots,v_{2n}\}$, where for each edge $e=\{v_i,v_j\}\in
E$ there are $n$ edges of the form $\{v_i,v_j,v_{n+k}\}, \
k=1,\dots,n$ in~$E'$. A 2-coloring for this hypergraph consists of
color classes $C_1=V$ and $C_2=\{v_{n+1},\dots,v_{2n}\}$. Any
maximum size independent set in this hypergraph consists of the
set of vertices $\{v_{n+1},\dots,v_{2n}\}$ joined with a maximum
size independent set~$S^*$ on~$G$. Therefore, any maximum size
independent set for~$H$ immediately induces a maximum size
independent set for~$G$, which is NP hard to obtain. Q.E.D.\\

The maximum size independent set problem in a hypergraph $H=(V,E)$
may be formulated as a Boolean linear programming problem
\begin{equation}\label{hyperBool}
\max\left\{ex : Ax \le b, x \in \{0,1\}^n\right\}
\end{equation}
with $m=|E|, n=|V|,$ $b_i=|e_i-1|, \ i=1,\dots,m$ and $a_{ij}=1$
iff $v_j \in e_i$, otherwise $a_{ij}=0$. In the special case
where~$H$ is 2-colorable, we can take $p^1$ and $p^2$ as the
indicator vectors for the color classes~$C_1$ and~$C_2$ of the
2-coloring. Then $D(p^1,p^2)=\{1,\dots,n\}$ and the OGTC for the
Boolean linear programming problem~(\ref{hyperBool}) is equivalent
to solving the maximum size independent set in a hypergraph~$H$
with a given 2-coloring, which leads to the following

\begin{corollary}
The OGTC for Boolean linear programming problem is NP-hard in the
strong sense even in the case where
all $|N_i|=3$, all $c_j=1$ and matrix~$A$ is Boolean.\\
\end{corollary}

Another example of an NP-hard OGTC problem is given by the set
covering problem, which may be considered as a special case
of~(\ref{goal})-(\ref{bools}):
\begin{equation}\label{setcover}
\min\left\{cx : Ax \ge e, \ x \in \{0,1\}^n\right\},
\end{equation}
$A$ is a Boolean $(m\times n)$-matrix. Let us assume the binary
representation of solutions by the vector $x$. Given an instance
of the set covering problem, one may construct a new instance with
a doubled set of columns in the matrix $A'=(AA)$ and a doubled
vector $c'={(c_1,\dots,c_n,c_1,\dots,c_n)}$. Then any instance of
the NP-hard set covering problem~(\ref{setcover}) is equivalent to
the OGTC for the set covering instance where the input consists of
$(m\times 2n)$-matrix $A'$, $2n$-vector~$c'$ and the parent
solutions $p^1,p^2,$ such that $p^1_j=1, p^2_{j}=0$ for
$j=1,\dots,n$ and $p^1_j=0, p^2_{j}=1$ for $j=n+1,\dots,2n$.

On the other hand, the OGTC problem for the set covering problem
is itself a set covering problem with reduced sets of variables
and constraints. So, the set covering problem is polynomially
equivalent to its OGTC problem.

The set covering problem may be efficiently transformed to the
simple plant location problem (see e.g. transformation~T3 \
in~\cite{KP83}) and this reduction meets the conditions of
Proposition~\ref{reduction0}, if the solution representation in
problem~(\ref{SPPLgoal})-(\ref{SPPLbool}) is given only by the
vector~$y$. Therefore, the OGTC for this version of the simple
plant location problem is NP-hard.

%------------------------------------------------------------------------
\section{Discussion \label{Discussion}}

As it was demonstrated above, even in the cases where the most
natural representation of solutions induces an NP-hard OGTC
problem, additional redundancy in the representation can make the
OGTC problem polynomially solvable. This seems to be a frequent
situation.

Another example of such case is the maximum 3-satisfiability
problem (MAX-3-SAT): given a set of~$M$ clauses, where each close
is a disjunction of three logical variables or their negations, it
is required to maximize the number of satisfied clauses~$f_{sat}$.
If a Boolean $N$-vector~$y$ encodes the assignment of logical
variables, then~$y$ is the most natural and compact representation
of solutions. Unfortunately, this encoding makes the OGTC problem
NP-hard (consider the parent solutions where $p^1_j+p^2_j=1, \
j=1,\dots,N$ -- then the OGTC becomes equivalent to the original
MAX-3-SAT problem, which is NP-hard).

Instead, we can move to a formulation of the MAX-3-SAT with a
graph-based representation, using a reduction from the MAX-3-SAT
to the maximum independent set problem, similar to the one
in~\cite{GJ}. In our reduction all vertices of the two-vertex
truth-setting components in the corresponding graph~$G=(V,E)$ are
given weight~$M$, the rest of the weights are equal to~1. On the
one hand, any truth assignment~$y$ for a MAX-3-SAT instance
defines an independent set in~$G$ with weight~$NM+f_{sat}(y)$ (the
mapping is described e.g. in~\cite{GJ}). On the other hand, any
independent set with weight~$NM+k, \ k\ge 0$ may be efficiently
mapped into a truth assignment~$y$ with $f_{sat}(y)\ge k$.
Obviously, all maximum-weight independent sets in~$G$ have a
weight at least~$NM$. So, solving the maximum-weight independent
set problem on~$G$ is equivalent to solving the original MAX-3-SAT
problem. We can consider only the independent sets of weight at
least~$NM$ as the feasible solutions to the MAX-3-SAT problem with
the described graph-based representation. Then the OGTC for this
problem is efficiently solvable by Proposition~\ref{reduction1}.
The general maximum satisfiability problem may be treated
analogously to MAX-3-SAT.

All of the polynomially solvable cases of the OGTC problem
considered above rely upon the efficient algorithms for the
maximum flow problem (or the maximum matching problem in the
unweighted case). However, the crossover operator initially was
introduced as a randomized operator. As a compromise approach one
can solve the optimized crossover problem approximately or solve
it optimally but only with some probability. Examples of the works
using this approach may be found in~\cite{BDE,Reeves94,HR96}.

In this paper we did not discuss the issues of GA convergence in
the case of optimized crossover. Due to fast localization of the
search process in such heuristics it is often important to provide
a sufficiently large initial population. Interesting techniques
that maintain the diversity of population by constructing the
second child, as different from the optimal offspring as possible,
can be found in~\cite{AOT97} and~\cite{BN98}. In fact, the general
schemes of the GAs and the procedures of parameter adaptation also
require a special consideration in the case of optimized
crossover.

%------------------------------------------------------------------------

\bibliographystyle{splncs}
\bibliography{paper14}

\end{document}